\documentclass{article}
\usepackage{arxiv}
\usepackage[utf8]{inputenc}
\usepackage[T1]{fontenc}
\usepackage{xcolor}
\usepackage{graphicx}
\usepackage{booktabs}
\usepackage{amsmath}
\usepackage{microtype}
\usepackage{float}
\usepackage{caption}
\usepackage{textcomp}
\usepackage{tcolorbox}
\usepackage{tikz}
\usetikzlibrary{positioning,arrows.meta,fit,backgrounds}
\usepackage{hyperref}
\renewcommand{\shorttitle}{LLM Performance on a Real, Double-Marked GCSE Benchmark}
\makeatletter
\renewcommand{\@maketitle}{%
  \vbox{%
    \hsize\textwidth
    \linewidth\hsize
    \vskip 0.3in
    \centering
    {\LARGE\bfseries \@title\par}
    \vskip 0.25in
    \def\And{%
      \end{tabular}\hfil\linebreak[0]\hfil%
      \begin{tabular}[t]{c}\bf\rule{\z@}{24\p@}\ignorespaces%
    }%
    \def\AND{%
      \end{tabular}\hfil\linebreak[4]\hfil%
      \begin{tabular}[t]{c}\bf\rule{\z@}{24\p@}\ignorespaces%
    }%
    \begin{tabular}[t]{c}\bf\rule{\z@}{24\p@}\@author\end{tabular}%
    \vskip 0.35in \@minus 0.1in
    \centerline{\@date}%
    \vskip 0.25in
  }%
}
\makeatother
\definecolor{accentnavy}{HTML}{14387F}
\definecolor{lightgrey}{HTML}{F5F5F5}
\definecolor{midgrey}{HTML}{CCCCCC}
\definecolor{softblue}{HTML}{E6EDF7}
\hypersetup{colorlinks=true,linkcolor=accentnavy,urlcolor=accentnavy,citecolor=accentnavy}
\newtcolorbox{examplebox}[1]{colback=white,colframe=midgrey,boxrule=0.5pt,arc=2pt,left=8pt,right=8pt,top=6pt,bottom=6pt,title={#1},coltitle=black,fonttitle=\bfseries\small,colbacktitle=lightgrey}
\title{LLM Performance on a Real, Double-Marked GCSE Benchmark}
\author{
  Malachy Fox \\ Medly AI \\ \texttt{malachy@medlyai.com}
  \And
  Kavi Samra \\ Medly AI \\ \texttt{kavi@medlyai.com}
  \And
  Paul Jung \\ Medly AI \\ \texttt{paul@medlyai.com}
}
\date{9 June 2026}
\begin{document}
\maketitle
\begin{abstract}
We introduce a dataset of 32{,}534 double-marked real student responses to GCSE mock exams (GCSEs are the UK's national exams, taken at age~16), spanning 328 questions across five subjects and including handwritten work. We test whether off-the-shelf large language models agree with examiners as closely as the two examiners agree with each other. We find that models overwhelmingly agree well with the examiner consensus across subjects, with the top performing models agreeing more closely with examiners than examiners agree with each other. Models achieve high scores for subjective tasks like English essay marking, as well as handling complex and messy handwritten Maths paper scripts. Agreement is uniform near the examiner line, and not massively discriminated by model size, providing cost-effective automated marking solutions.
\end{abstract}

\section{Introduction}

Exam boards and schools require reliable marking \cite{ofqual2018}, but manual assessment imposes a heavy workload on teachers \cite{dfe2019}. Automated essay scoring has a long history, progressing from early feature-based and neural systems \cite{shermis2013,asap2012,taghipour2016,dong2017} to recent large language model approaches \cite{xiao2024,aimecon2025,mansour2024,trates2025}, including subject-specific scoring such as science assessment \cite{latif2024}. Early results were mixed: prompted off-the-shelf models fell short of task-specific systems on standard essay benchmarks \cite{mansour2024}, and later work recovered ground by using the model to generate rubric-grounded features rather than to score directly \cite{trates2025}. Most school examinations, though, span multiple subjects and include handwritten equations, short text answers, and diagrams, which automated systems find hard to mark reliably \cite{kortemeyer2024,caraeni2024}; even on born-digital multimodal mathematics, leading models still trail humans \cite{mathvista2024}. We ask whether commercially available LLMs, run under a generic prompt at minimum reasoning effort, are already a reliable second marker across these response types, and where they fail.

We present a benchmark of 32{,}534 double-marked GCSE mock responses across 328 questions in English Language, Maths, Biology, Chemistry, and Physics. GCSEs are the national subject examinations taken by students in England, Wales and Northern Ireland at age $\sim 16$, marked by qualified examiners against published mark schemes; the ``Higher'' tier labelled in the tables and worked examples is the more demanding of the two GCSE tiers in Maths and the sciences. To our knowledge it is the first publicly described double-marked, multi-subject GCSE marking benchmark to include handwritten student work alongside typed text. We evaluate models by their average agreement with each examiner, measured against the agreement between the two examiners themselves. This isolates whether a model marks as consistently as a human marker.

We report agreement subject by subject, and investigate how models are differentiated by bias when marking English essays.

\section{The Dataset}

The dataset comprises 32,534 student responses across 328 questions in five subjects (English Language, Maths, Biology, Chemistry, and Physics), sampled to span the full attainment range. Each answer was independently marked by two qualified examiners. Answers combine typed text, on-screen text boxes, and freehand handwriting and drawings captured as strokes, so a marker must be multimodal. The two worked examples below show the range: a typed English essay, and a handwritten Maths item. Table~\ref{tab:handwriting} gives the share of handwritten input per subject.

\begin{examplebox}{English Language (creative writing, 40 marks)}
\textbf{Prompt to the model.} \textit{``Mark the student's answer using the mark scheme provided. Respond with JSON only, in the form \{"mark": <integer>\}.''} This is sent with the question, mark scheme and student answer below.

\textbf{Question.} Your local newspaper is running a creative writing competition. \emph{Either:} write a description of a storm at sea as suggested by a picture; \emph{or:} write a story about an unexpected visitor.

\textbf{Mark scheme.} Two assessment objectives, marked separately on best-fit level descriptors. \textbf{AO5, content and organisation} (24 marks): communicate clearly and imaginatively, matching tone, style and register to purpose and audience, with coherent structure. \textbf{AO6, technical accuracy} (16 marks): a range of vocabulary and sentence structures, accurate spelling and punctuation.

\textbf{Student answer.} It was a quiet room, I was doing my regular daily routines as normality invaded me\ldots{} Then I heard it, a loud knock that invaded my conscience; the sound didn't belong and my chest was beating faster than ever. \textnormal{[\ldots]} For a second, I didn't recognise him: he looked thinner somehow, sharper around the edges like time had stripped something away. \textnormal{[\ldots]} He stepped into the dark and whispered, ``some consequences follow you home.''

\textit{Examiner marks: 19/40 (both examiners; AO5 + AO6 combined).}
\end{examplebox}

\begin{examplebox}{Maths Higher (highest common factor, 2 marks)}
\textbf{Prompt to the model.} \textit{``Mark the student's answer using the mark scheme provided. Respond with JSON only, in the form \{"mark": <integer>\}.''} This is sent with the question, mark scheme and student answer below. The rendered canvas image is attached alongside it.

\textbf{Question.} Find the highest common factor (HCF) of 84 and 126.

\textbf{Mark scheme.} \textbf{M1} for a correct method, e.g.\ the prime factorisation of either number ($84 = 2 \times 2 \times 3 \times 7$ or $126 = 2 \times 3 \times 3 \times 7$). \textbf{A1} for $42$ (or $2 \times 3 \times 7$).

\textbf{Student answer.} The student worked by hand on the canvas: trial division and factor trees to find the prime factors of each number, the shared factors combined, and the answer $42$ circled (see below).

\begin{center}\includegraphics[width=0.62\linewidth]{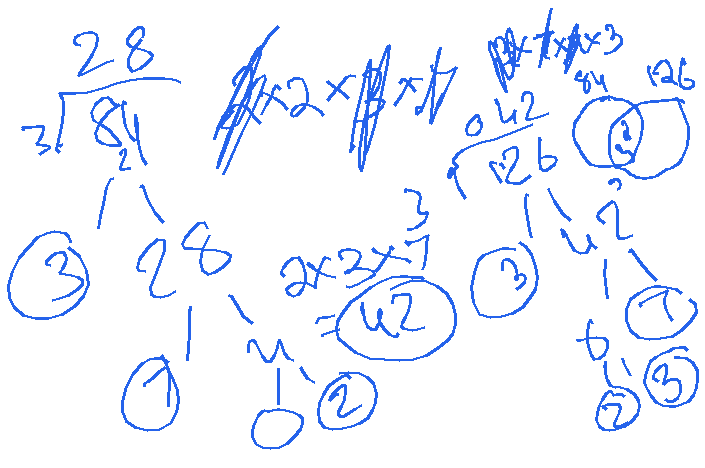}\end{center}

\textit{Examiner marks: 2/2 (both examiners). AI mark: 2/2.}
\end{examplebox}

\begin{table}[H]
\centering
\caption{Share of responses containing handwritten input (on-screen handwriting captured as strokes).}
\label{tab:handwriting}
\begin{tabular}{lrrr}
\toprule
\textbf{Subject} & \textbf{Total} & \textbf{Handwritten} & \textbf{HW \%} \\
\midrule
        English Language & 899 & 0 & 0.0\% \\
        Maths Higher & 6,013 & 2,650 & 44.1\% \\
        Biology Higher & 8,207 & 336 & 4.1\% \\
        Chemistry Higher & 8,808 & 873 & 9.9\% \\
        Physics Higher & 8,607 & 1,123 & 13.0\% \\
        \textbf{Total} & \textbf{32,534} & \textbf{4,982} & \textbf{15.3\%} \\
\bottomrule
\end{tabular}
\end{table}

\section{Method}

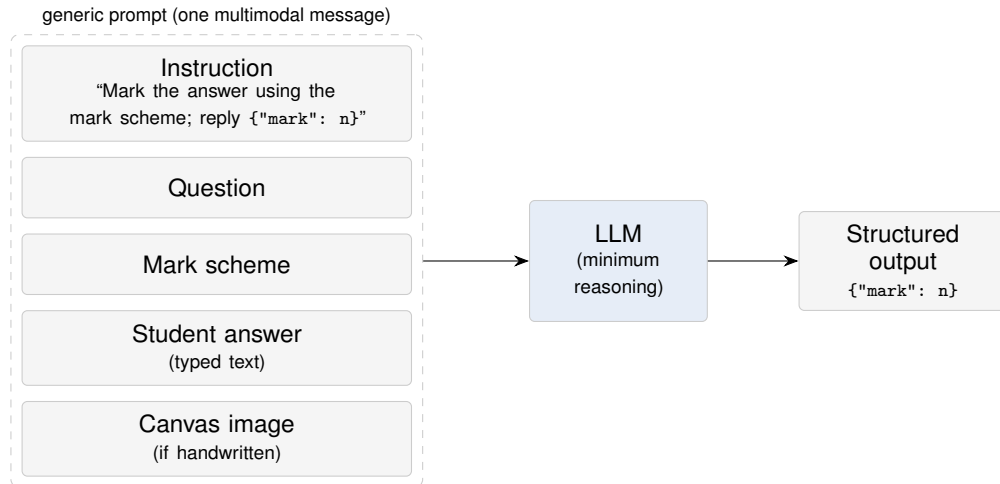
\begin{figure}[H]
\centering
\begin{tikzpicture}[
    font=\sffamily\small,
    box/.style={draw=midgrey,rounded corners=2pt,align=center,inner sep=5pt,minimum height=8mm},
    inp/.style={box,fill=lightgrey,text width=48mm},
    model/.style={box,fill=softblue,text width=20mm,minimum height=16mm},
    outbox/.style={box,fill=lightgrey,text width=24mm},
    >={Stealth[length=2mm]}]
  \node[inp] (instr) {Instruction\\\scriptsize``Mark the answer using the\\mark scheme; reply {\ttfamily\{"mark": n\}}''};
  \node[inp,below=2mm of instr] (q)  {Question};
  \node[inp,below=2mm of q] (ms) {Mark scheme};
  \node[inp,below=2mm of ms] (ans) {Student answer\\\scriptsize(typed text)};
  \node[inp,below=2mm of ans] (img) {Canvas image\\\scriptsize(if handwritten)};
  \begin{scope}[on background layer]
    \node[draw=midgrey,dashed,rounded corners=3pt,fit=(instr)(q)(ms)(ans)(img),inner sep=4pt,label=above:{\scriptsize generic prompt (one multimodal message)}] (msg) {};
  \end{scope}
  \node[model,right=14mm of msg] (m) {LLM\\[1pt]\scriptsize(minimum\\reasoning)};
  \node[outbox,right=12mm of m] (mark) {Structured output\\\ttfamily\scriptsize\{"mark": n\}};
  \draw[->] (msg.east) -- (m.west);
  \draw[->] (m.east) -- (mark.west);
\end{tikzpicture}
\caption{The generic marking prompt: a single multimodal message (instruction, question, mark scheme, student answer, and any canvas image) returns a structured integer mark.}
\label{fig:prompt}
\end{figure}

Each response was marked with a generic prompt: the question, mark scheme, and student answer, with a one-line instruction to mark it out of the maximum (Figure~\ref{fig:prompt}). For handwritten work the rendered canvas image is attached, so the model must read the handwriting itself. The model is constrained to structured output, a JSON schema with a single integer \texttt{mark} field. Models were run at minimum reasoning effort.\footnote{``Minimum'' is the lowest reasoning setting each model exposes: this disables reasoning entirely for every model except Gemini 3.1 Pro, whose lowest available setting is `low' rather than `off'.}

Marks are ordinal, so agreement is measured with Quadratic Weighted Kappa (QWK) \cite{cohen1968}, computed per question and then averaged across questions \emph{weighted by each question's maximum mark}. This matches how a student's grade is formed, a sum of marks in which a 6-mark question counts six times a 1-mark one, and it damps the noise of the many near-binary low-mark items. QWK runs from 0 (chance agreement) to 1 (identical marks). On the standard scale \cite{landis1977} 0.21--0.40 is fair, 0.41--0.60 moderate, 0.61--0.80 substantial, and 0.81--1.00 almost perfect, so a QWK around 0.8 already indicates close agreement between two raters.

We report the average of examiner-model agreement against examiner-examiner agreement. For a model the mean agreement with the two examiners is
\begin{equation*}
R_{A} = \tfrac{1}{2}\left[\mathrm{QWK}(\mathrm{AI},E_1) + \mathrm{QWK}(\mathrm{AI},E_2)\right],
\end{equation*}
the mean QWK between the model and each examiner. The reference is the agreement between the two examiners, $R_{H} = \mathrm{QWK}(E_1,E_2)$, where $E_1,E_2$ are the two examiners. So $R_A$ is the model's mean agreement with the two examiners, and $R_H$ is how closely the two examiners agree with each other. An automated marker should land on the mark the two examiners would settle on. Each $R_A$ and $R_H$ carry a 95\% confidence interval from a cluster-over-questions bootstrap (2{,}000 resamples). We report the difference $\Delta = R_A - R_H$ with its interval (Table~\ref{tab:top}).

For essays we also report \textbf{signed marking bias}: mean$(\text{predicted} - \text{consensus})$ as a fraction of the maximum mark, where consensus is the two-examiner mean (positive = lenient, negative = harsh).

\section{Agreement with the Examiners}\label{sec:bench}

\begin{figure}[H]
\centering
\includegraphics[width=0.74\textwidth]{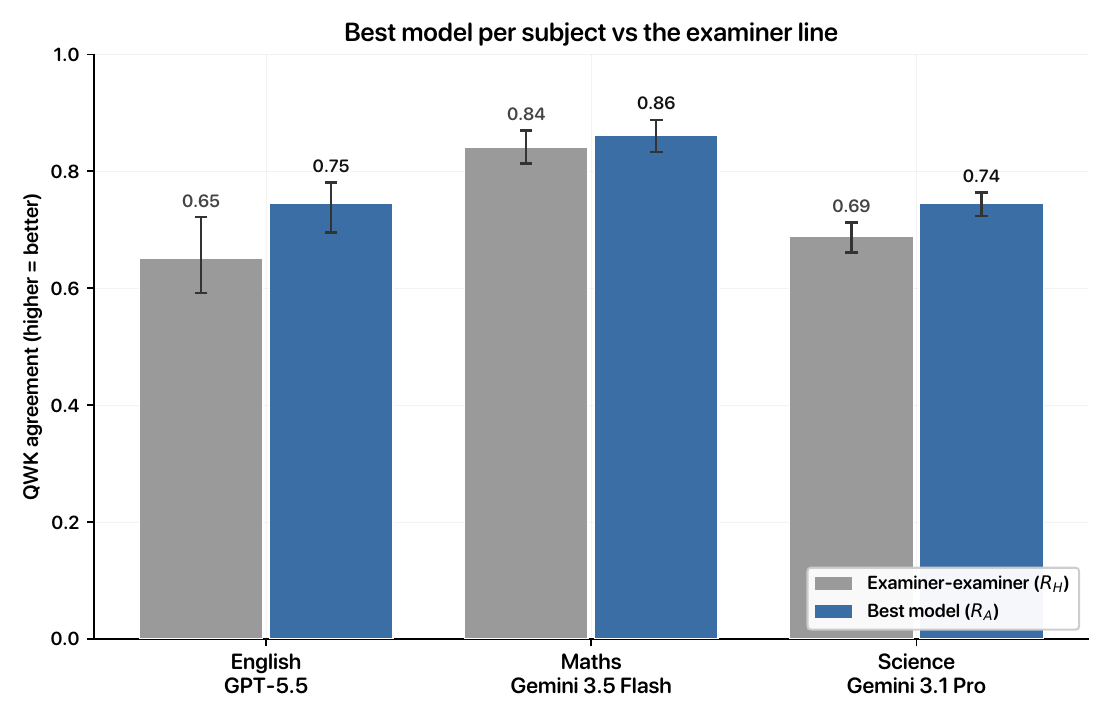}
\caption{Best model per subject ($R_A$, coloured) against the examiner-examiner agreement ($R_H$, grey). A coloured bar reaching the grey marks as consistently as a second examiner.}
\label{fig:pairwise}
\end{figure}

We investigate whether a model agrees with two examiners as closely as the two examiners agree with each other. Figure~\ref{fig:pairwise} shows each best-in-subject model's mean agreement, $R_A$, against the agreement between the two examiners, $R_H$. Table~\ref{tab:top} gives the per-subject values and the difference $\Delta = R_A - R_H$ with its 95\% confidence interval.

Across all subjects, the leading models agree with an examiner more closely than the two examiners agree with each other. English shows the best performance, with models of all sizes reaching examiner agreement and top performing models greatly exceeding it. In Maths, although the examiner agreement is very high at around 0.84, the top performing model achieves a delta of +0.02. Science also shows examiner-level marking ability, with a highest delta of +0.06.

Differences between models are small and fall within the confidence intervals, and model size does not predict agreement: small models mark as consistently as large ones. The leading models per subject are GPT-5.5 in English Language, Gemini 3.5 Flash in Maths, and Gemini 3.1 Pro across the pooled sciences, with more cost-effective models scoring very similarly. The sections that follow break these results down.

\begin{table}[H]
\centering
\small
\caption{Numeric values for Figure~\ref{fig:pairwise}: the best model's agreement $R_A$, the examiner line $R_H$, and $\Delta = R_A - R_H$ with its 95\% CI. $\Delta > 0$ (CI clear of zero) means the model marks closer to the examiners than they do to each other.}
\label{tab:top}
\begin{tabular}{llcccc}
\toprule
\textbf{Subject} & \textbf{Best model} & \textbf{$R_A$} & \textbf{$R_H$} & \textbf{$\Delta$} & \textbf{95\% CI} \\
\midrule
        English & GPT-5.5 & 0.75 & 0.65 & $+0.10$ & $[+0.04,\,+0.14]$ \\
        Maths & Gemini 3.5 Flash & 0.86 & 0.84 & $+0.02$ & $[+0.00,\,+0.04]$ \\
        Science & Gemini 3.1 Pro & 0.74 & 0.69 & $+0.06$ & $[+0.04,\,+0.07]$ \\
\bottomrule
\end{tabular}
\end{table}

\section{Essay Marking}\label{sec:essay}

\begin{figure}[H]
\centering
\includegraphics[width=0.96\textwidth]{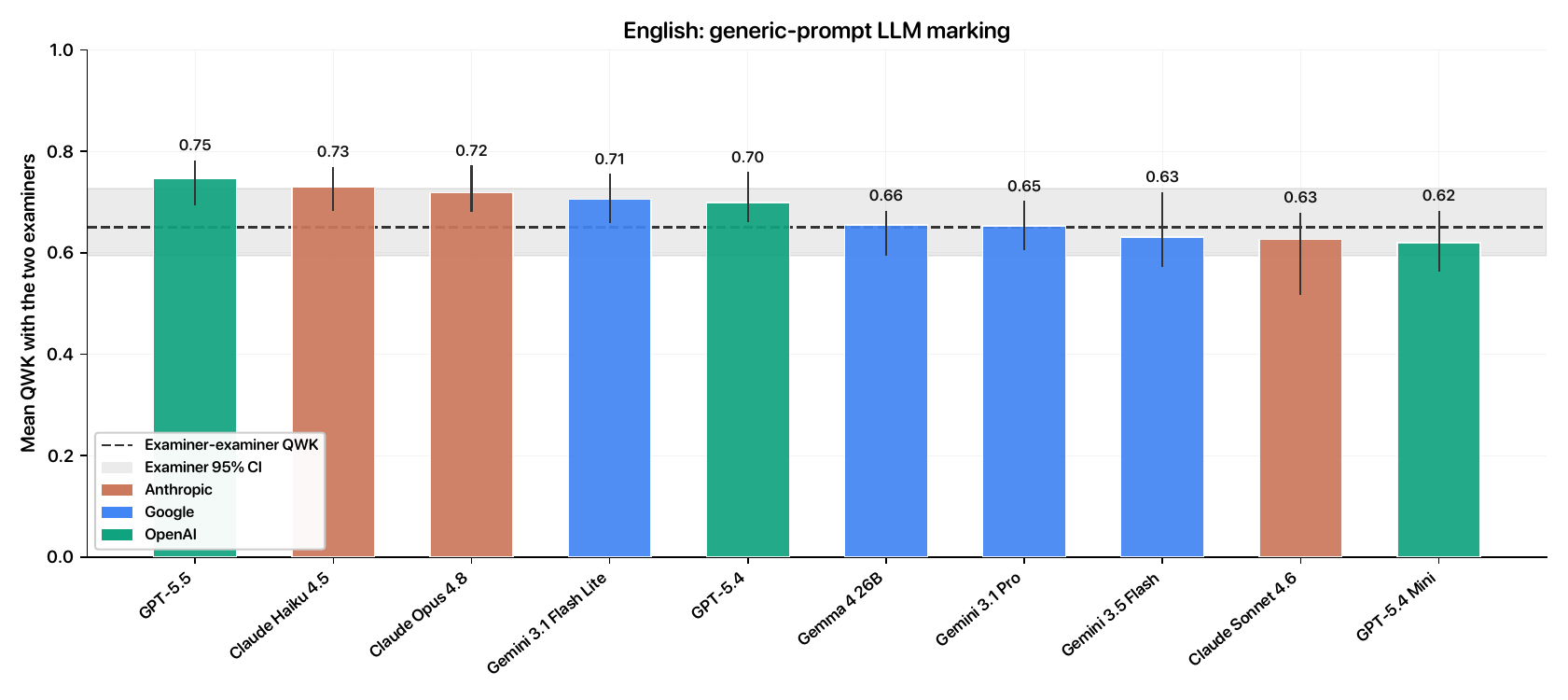}
\caption{English: each model's mean QWK with the two examiners (higher = better), best first. Dashed line and band are the examiner-examiner QWK and its 95\% CI; a bar reaching the band marks as consistently as a second examiner. Figures~\ref{fig:lead-maths} and~\ref{fig:lead-science} share this layout.}
\label{fig:lead-english}
\end{figure}

Essays are the most subjective marking task. Examiner-examiner agreement is much lower than the STEM subject agreement rates. Additionally, the English assessment rests on only eight questions, so the per-subject QWK in Figure~\ref{fig:lead-english} is comparatively noisy. Prior work applying GPT-4 to second-language essay assessment found significant correlations with human scores on a single annotated dataset \cite{banno2024}; here we mark real GCSE essays against a double-marked examiner baseline, asking not whether the model correlates with humans but how it sits relative to the spread between two markers.

\begin{figure}[H]
\centering
\includegraphics[width=0.82\textwidth]{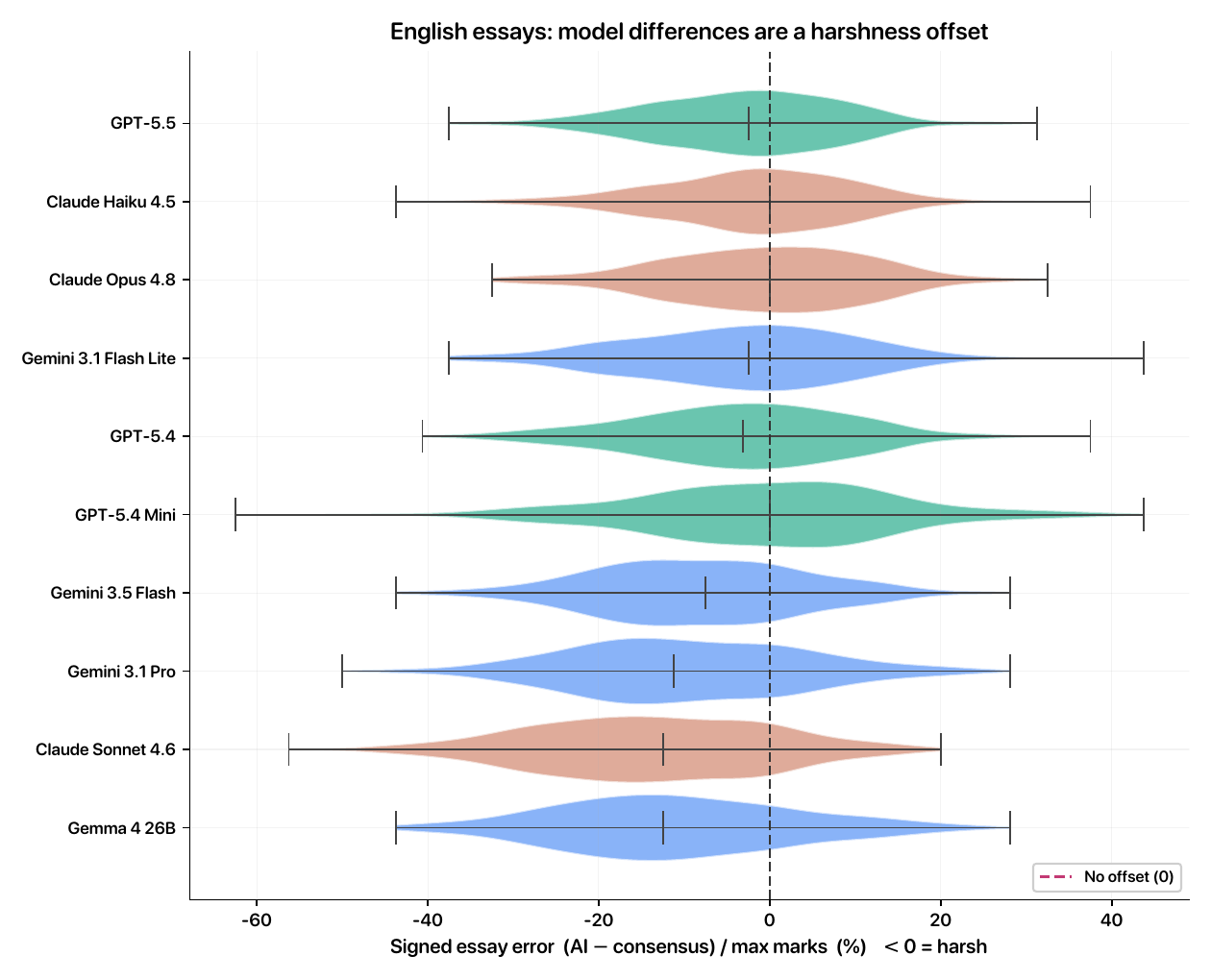}
\caption{English essays, signed error per model (AI $-$ consensus, \% of max mark), most accurate at the top. Dashed line is neutral; left is harsh, right lenient.}
\label{fig:essay-harsh}
\end{figure}

The models still agree with an examiner more closely than the two examiners agree with each other ($R_A > R_H$). The spread of errors is similar across models, and does not correlate with model size/cost. One distinguishing factor, however, is marking bias, shown in Figure~\ref{fig:essay-harsh}. The difference is a harshness offset. Claude Opus 4.8, Claude Haiku 4.5, and GPT-5.5 are the most neutral, while Claude Sonnet 4.6 and Gemma 4 26B are the harshest.

\section{Science and Maths Marking}\label{sec:stem}

The majority of tested models were found to mark the sciences and Maths as consistently as a second examiner, with Maths showing larger gaps in performance between models. Top models in Science achieve positive deltas approaching +0.06, showing an improvement in consistency over examiners.

\begin{figure}[H]
\centering
\includegraphics[width=0.96\textwidth]{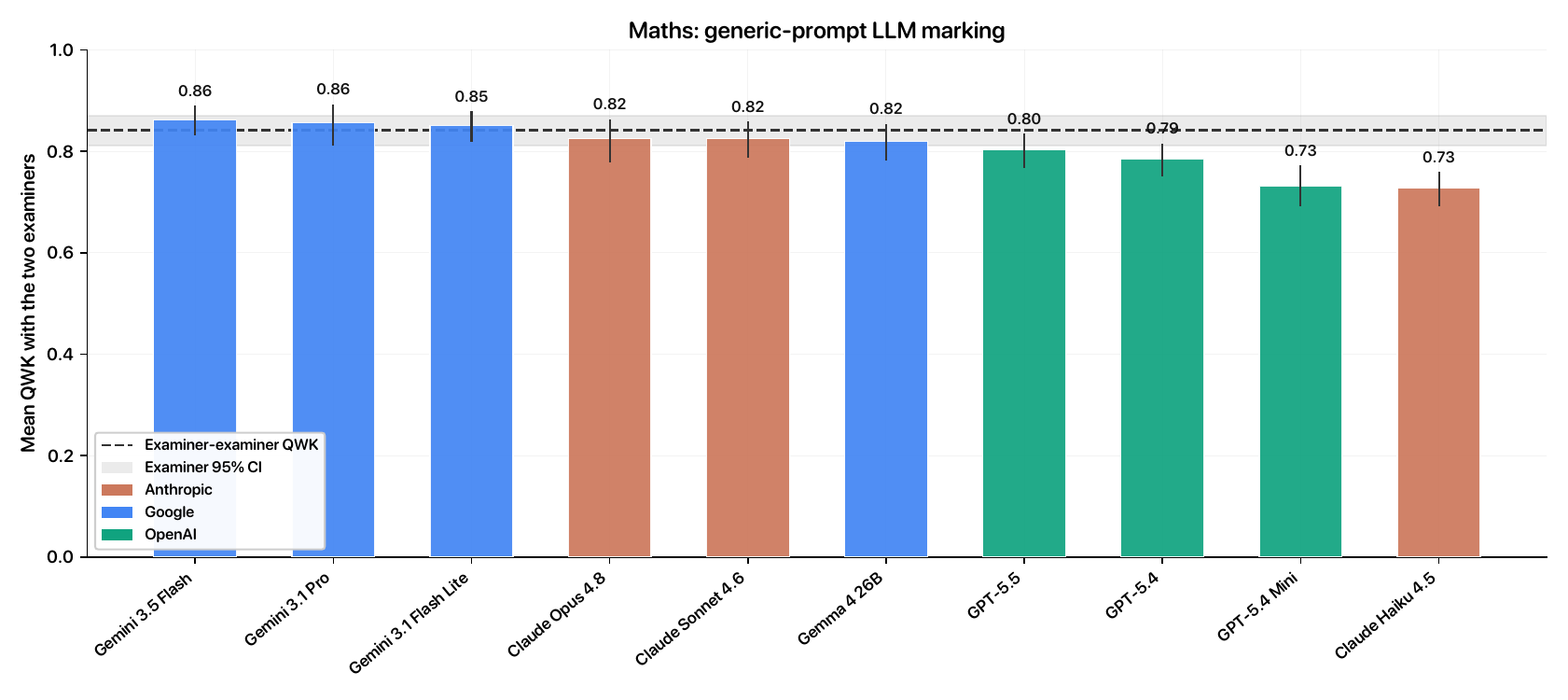}
\caption{Maths: model mean QWK with the two examiners against the examiner-examiner line (cf.\ Figure~\ref{fig:lead-english}).}
\label{fig:lead-maths}
\end{figure}

In Maths the examiner line is high, an $R_H$ of 0.84, because the two examiners agree very tightly. Google's Gemini models seem to be more aligned with the expectation of the GCSE marking criteria than OpenAI's models, perhaps an artefact of their training data, appearing to be a stronger indicator than model size. (Figure~\ref{fig:lead-maths}).

\begin{figure}[H]
\centering
\includegraphics[width=0.96\textwidth]{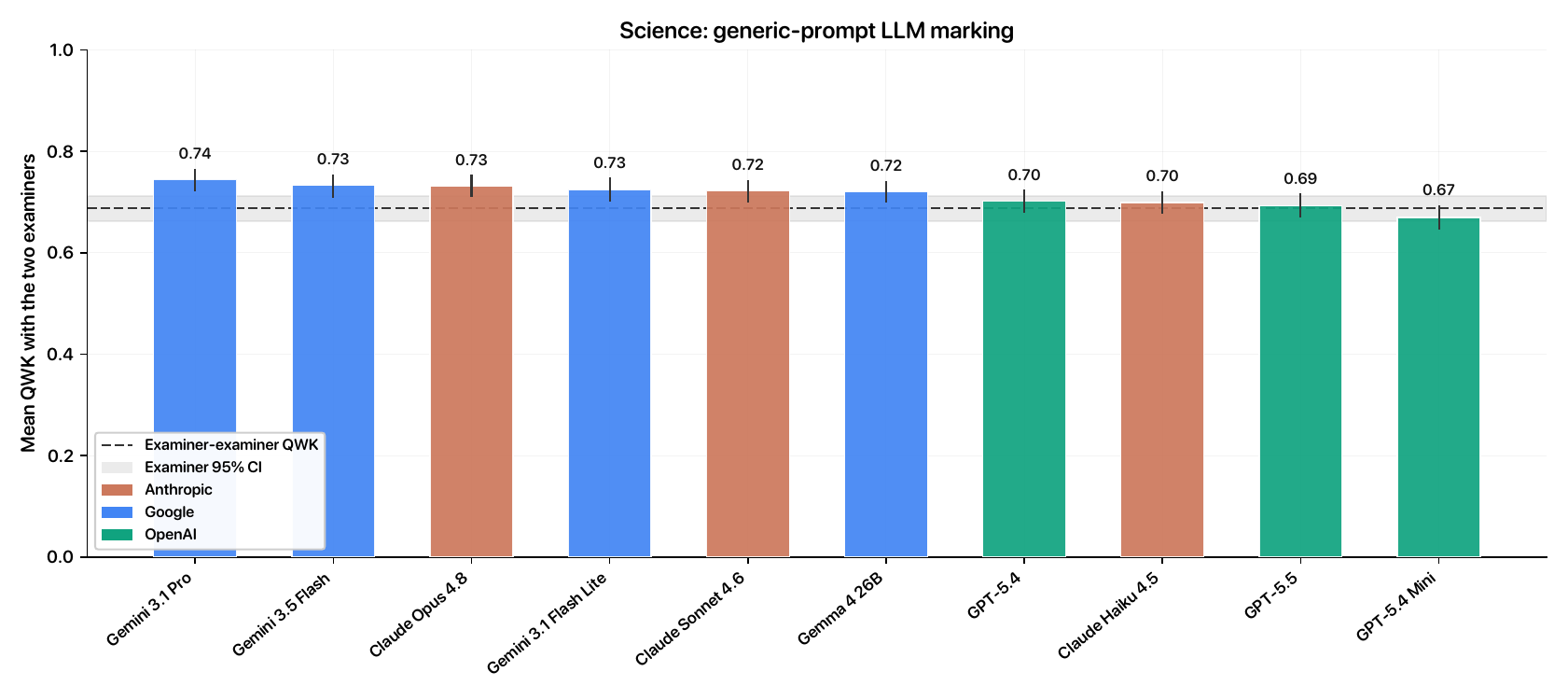}
\caption{Science (Biology, Chemistry, Physics combined): model mean QWK with the two examiners against the examiner-examiner line (cf.\ Figure~\ref{fig:lead-english}).}
\label{fig:lead-science}
\end{figure}

\section{Cost and Model Selection}\label{sec:cost}

\begin{figure}[H]
\centering
\includegraphics[width=0.95\textwidth]{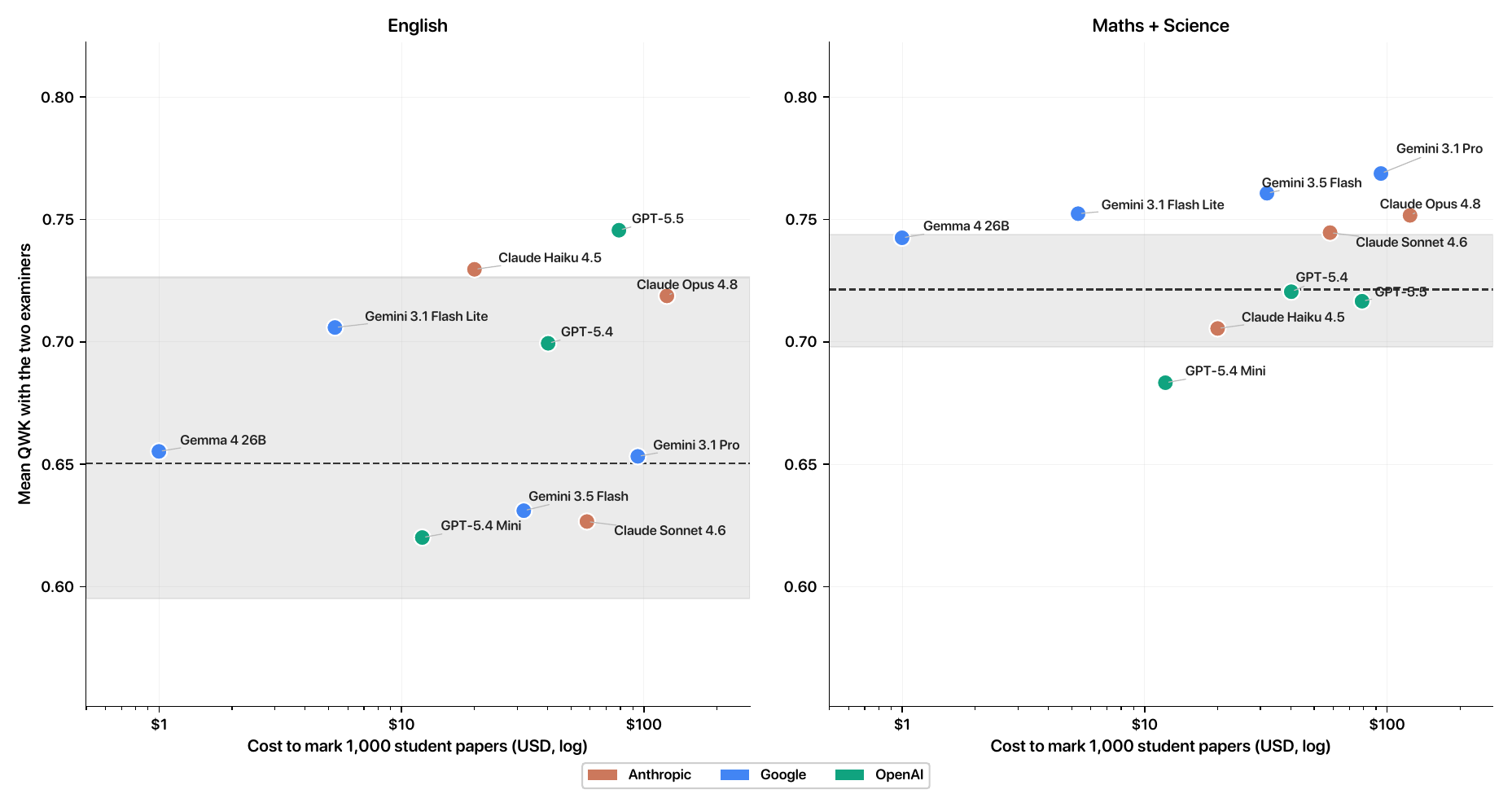}
\caption{Mean QWK vs cost to mark 1{,}000 papers (USD list price, log scale), for English and Maths+Science. Each point is a model; top-left is best (high agreement, low cost). Dashed line and band are the examiner line and its 95\% CI.}
\label{fig:cost}
\end{figure}

Marking 1{,}000 student papers costs between \$1 to \$120 at list price, as shown in Figure~\ref{fig:cost}. Cost correlates only weakly with agreement; several cheaper models such as Claude Haiku 4.5 for English, and Gemini 3.1 Flash Lite for Science/Maths, achieve better performance than Claude Opus 4.8 for a fraction of the cost.

\section{Discussion and Conclusion}

This benchmark shows that the LLMs available today, run under a single generic prompt at minimum reasoning effort, work as a reliable second marker. Additionally, there is scope for improved performance under improved prompting or even fine-tuning. A double-marked, multi-subject dataset that includes handwriting tests this under realistic conditions, overcoming limitations of other benchmarks.

We note several limitations. The dataset comes from Medly's own mock examinations, so the question types mimic official GCSE exam papers rather than using them directly. We also evaluate models only at minimum reasoning effort, to explore a best-case cost and latency scenario for each model.

Finally, the two-examiner consensus carries a variance caveat. With only two raters, between-examiner variance is one degree of freedom. It does not mean models are compared to the broader examiner population. Characterising the entire examiner population would require more than two raters.

\section*{Data Availability}

A public companion subset of the benchmark is released at \url{https://github.com/medlyai/medly-marking-benchmark}.

\end{document}